%% file: main.tex
\documentclass[runningheads]{llncs}

\usepackage{eccv}

\input{preamble}

\begin{document}
\maketitle
\definecolor{violetarrow}{RGB}{148,0,211}

\begin{abstract}

We present \ours, a unified vision–language foundation model capable of both image understanding and image generation within a single autoregressive framework. Unlike existing vision models that depend on separate modules for perception and synthesis, \ours adopts a fully integrated architecture that enforces cycle-consistent learning through image→layout→image and layout→image→layout generation loops. This unified formulation introduces two key advantages: {\bf introspection}, enabling the model to reason about its own generations, and {\bf data efficiency}, allowing self-improvement via synthetic supervision under a reinforcement learning objective guided by cycle consistency. Extensive experiments show that \ours achieves significant gains across diverse image understanding and generation benchmarks, highlighting the potential of unified vision–language foundation models.
\end{abstract} 
\begin{figure*}[t]
\centering

\includegraphics[width=\linewidth]{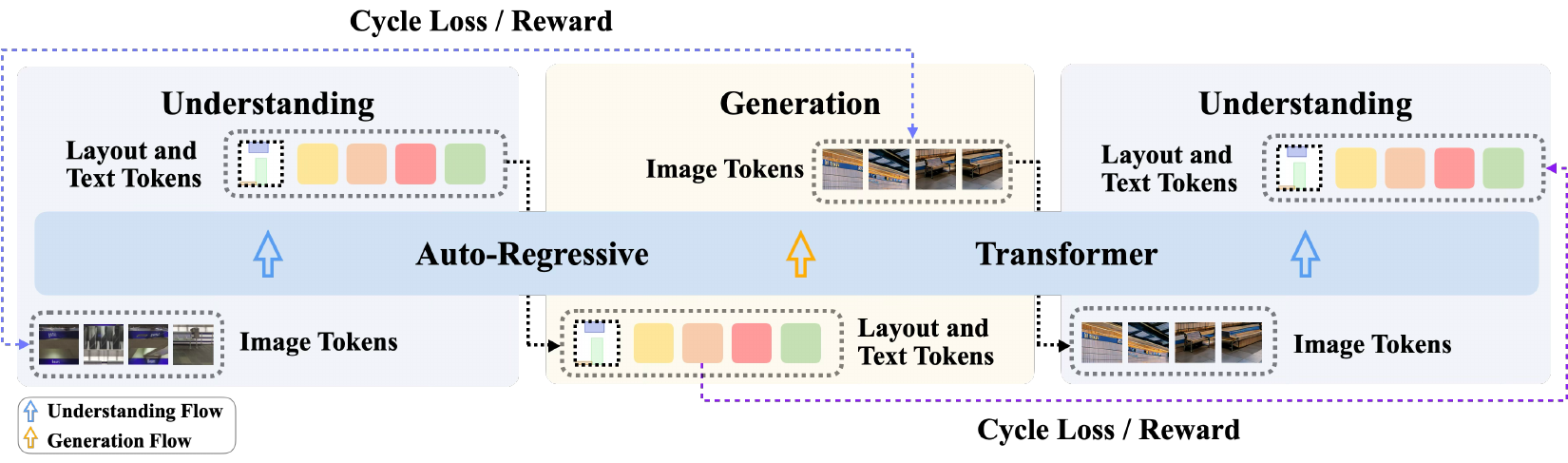}

\caption{\small
\textbf{Overview of the CyCLeGen framework. }
\ours employs a unified autoregressive transformer that jointly performs 
layout understanding (image → layout) and layout-conditioned image generation 
(layout → image) within a shared token space. 
Given image tokens, the model predicts layout and text tokens through the 
understanding branch; conversely, conditioned on layout and text tokens, 
the generation branch produces image tokens. 
Training enforces bidirectional cycle consistency across the two directions, 
forming image → layout → image and layout → image → layout loops. 
Cycle rewards/losses encourage the generated outputs in one direction to remain 
structurally and semantically recoverable in the reverse direction, aligning 
visual understanding with controllable image generation.
}
\label{fig:main_pipeline}
\end{figure*}
\section{Introduction}\label{sec:intro}
Unified understanding-generation models have advanced rapidly in recent years \cite{zhang2025unified}, revealing that a single transformer architecture can support a wide range of perception and visual synthesis tasks.
Large-scale models, such as Show-O \cite{xie2024showo}, Transfusion \cite{zhou2024transfusion}, and the Janus series \cite{chen2025janus, wu2025janus}, demonstrate that by jointly optimizing generation
and understanding with a crafted unified architecture can perform diverse tasks---such as detection, captioning, and text-to-image generation---without requiring separate, task-specific modules.

However, a crucial question remains unanswered - \textbf{Do unified models truly benefit from jointly learning visual \emph{understanding} and \emph{generation}?}
While intuition suggests a synergistic relationship: understanding should ground generation, and generation should enrich visual representations, in practice, joint training is often unstable.
We observe a significant \textbf{task conflict}: optimizing the model for visual understanding frequently degrades visual generative fidelity, while optimizing for high-fidelity generation can, in turn, weaken the visual understanding performance.
This mutual interference poses a central challenge for unified vision-language training:
\emph{How can we simultaneously improve both understanding and generation capabilities without one task compromising the other?}

Most existing unified models \cite{deng2025emerging, he2025plangen, chen2025janus} address multi-task learning through 
\emph{interleaved training}, where understanding and generation samples are randomly mixed within the same optimization stream. Although simple and scalable, this strategy treats tasks as independent objectives that merely share parameters, without explicitly modeling their structural dependency. In contrast, we explicitly couple the two inverse tasks through cycle-consistent supervision. Rather than mixing objectives, we  \emph{constrain} them: generation must be re-parseable by the understanding branch, and understanding must support faithful reconstruction. This structural coupling transforms task competition into mutual regularization, leading to stable co-improvement instead of gradient interference. 

To study this challenge in a controlled, interpretable, and structurally rigorous setting, we focus on a pair of tasks that are exact inverses of one another:  
\textbf{layout understanding} (image $\rightarrow$ layout) and \textbf{layout-to-image generation} (layout $\rightarrow$ image).  
This pair provides an ideal \emph{proxy instantiation} of the more general text $\rightarrow$ image $\rightarrow$ text cycle for several reasons:   
(1) Layout offers a compact and interpretable intermediate representation.  
(2) The two directions form a mathematically clean inverse mapping.
(3) Enforcing agreement between them naturally leads to more controllable and structurally aligned image generation.  
Thus, layout–image tasks allow us to precisely analyze—and ultimately overcome—the interference behaviors inherent to unified multimodal models.

With this in mind, we present \ours: a \textbf{Cy}cle-\textbf{C}onsistent \textbf{L}ayout-based unified autoregressive framework that brides layout understanding and layout-to-image generation through an explicit cycle consistency constraint.  
The model learns a stable mapping between images and layouts, enabling consistent reconstruction across both directions, i.e. image $\rightarrow$ layout $\rightarrow$ image and layout $\rightarrow$ image $\rightarrow$ layout. 

To further address optimization conflicts, we introduce \textbf{CycleGRPO}, a \textbf{Bidirectional Cycle-consistent Reinforcement Learning} strategy that resolves the asymmetric optimization pressures between visual understanding and generation in interleaved unified multi-modal modeling. 
Unlike prior reinforcement learning approaches that optimize either image generation\cite{wei2025skywork} or visual understanding\cite{liu2025visual} in isolation, ur method jointly aligns the two directions through cycle-based optimization.
Specifically, we alternate between two complementary directions:
In the \textit{understanding$\rightarrow$generation} direction, the model updates the understanding branch using IoU and CLIP rewards, encouraging structurally accurate layout predictions that support high-fidelity image synthesis. 
In the reverse \textit{generation$\rightarrow$understanding} direction, the model updates the generation branch using IoU and HPSv2 rewards, improving perceptual fidelity, texture realism, and layout controllability. 
By interleaving these updates, the model avoids reward interference while enforcing a cycle constraint that aligns generation and understanding within a unified representation.

\paragraph{Advantages.}
Our framework offers several key benefits:

\noindent (1) \textbf{Introspection.}  
Because understanding and generation are implemented within a single autoregressive model, \ours can \textbf{self-evaluate} its generations by re-parsing them through its own understanding branch.  
This intrinsic feedback provides a natural learning signal,

 and offers a scalable mechanism for synergistic refinement.

\noindent (2) \textbf{Data efficiency.}  
Instead of millions of paired understanding–generation examples, \ours achieves notable performance gains with only 
8k RL samples.  
Cycle consistency enables the model to train on its own predictions, effectively turning the layout $\rightarrow$ image $\rightarrow$ layout loop into a form of synthetic supervision that reduces dependency on large annotated datasets.

\paragraph{Contributions.}
Our main contributions are as follows:
\begin{itemize}
    \item To the best of our knowledge, we are the first to introduce cycle-consistency into an unified autoregressive framework to attain {\bf introspection} and {\bf data efficiency} capabilities, jointly supervising both layout understanding and layout-conditioned generation.
    \item We propose \textbf{CycleGRPO}, a bidirectional cycle-constrained reinforcement learning strategy that jointly optimizes understanding and generation with complementary geometric and perceptual rewards.

    \item Extensive experiments show that our unified approach substantially improves structural fidelity, perceptual realism, and controllable generation, outperforming \textbf{converged} unified baselines.

\end{itemize}

\section{Related Work}\label{sec:formatting}

\paragraph{Unified Model}
Large vision-language models (LVLMs)~\cite{grattafiori2024llama, hui2024qwen2.5-paper,alayrac2022flamingo,achiam2023gpt4-turbo,Brown2020GPT3} have rapidly advanced toward unified multimodal understanding and generation. Recent architectures such as Show-o~\cite{xie2024showo}, Transfusion~\cite{zhou2024transfusion}, MonoFormer~\cite{zhao2024monoformer}, Emu3~\cite{wang2024emu3}, Janus/Janus-Pro~\cite{wu2025janus,chen2025janus}, Bagel~\cite{deng2025emerging} and Chameleon~\cite{team2024chameleon} attempt to model perception and synthesis within one transformer. 
However, these unified multimodal models mainly focus on text–image domain and lack \emph{cycle-consistent learning} between understanding and generation. Our work differs by introducing a unified framework explicitly linking layout understanding and layout-to-image generation through bidirectional cycle-consistent supervision.  

\paragraph{Layout Understanding and Layout-to-Image Generation}
Layout-to-image generation has been explored through both training-free approaches~\cite{chen2024training_cag,xie2023boxdiff} and training-based diffusion models~\cite{li2023gligen,cheng2024hico,wang2024instancediffusion,zhou2024migc,zhang2024creatilayout, Srivastava_2025_ICCV}. Layout is also widely used as a proxy for general generation scenario\cite{Zeng_2025_ICCV, Wang_2024_CVPR,chen2025cvp}. 
While these methods improve spatial controllability, they operate solely in the \emph{generation} direction and do not learn how layouts map back from images.  
PlanGen\cite{he2025plangen} is among the first attempts to unify layout understanding and generation in an autoregressive model, but its formulation is still limited: understanding and generation remain loosely coupled, lacking mutual constraints, and the supervision on understanding is sparse.  
Our method addresses these limitations by enforcing \emph{cycle-consistency} between layout to image and image to layout, enabling stronger structural correctness and verifiable feedback between the two tasks.

\paragraph{Reinforcement Learning for LVLMs}
Reinforcement learning has recently become central to advancing reasoning in large language models, with progress in math~\cite{grpo,yang2024qwen2math,ying2024internlmmath}, coding~\cite{hui2024qwen2.5-paper,jiao2024preferencecode}, and pure-RL systems such as DeepSeek-R1-Zero~\cite{guo2025deepseek}.  
In vision-language settings, RL has primarily targeted hallucination mitigation and preference alignment~\cite{hadpo,sun2023aligning,yu2024rlhf}, with little focus on visual reasoning or structured perception. 
Prior works also lack \emph{verifiable rewards} that connect visual understanding and generation.  
We introduce a GRPO-based~\cite{grpo} reinforcement fine-tuning pipeline that provides cycle-aware, automatically verifiable rewards for both layout understanding and layout-conditioned generation, significantly improving structured perception under limited supervision.

\paragraph{Introspective Learning}
The framework of energy-based generative adversarial learning was first proposed in \cite{tu2007learning}, with a single unified model being both generative and discriminative. This approach was subsequently evolved into Introspective Neural Networks (INN) \cite{jin2017introspective,lazarow2017introspective,lee2018wasserstein}. While INNs primarily target binary and multi-class image synthesis and classification, \ours functions as a foundation model, offering significantly broader scope and enhanced capabilities.

\section{Method}

\subsection{Overview}

\ours aims to improve the cycle-consistent alignment between the layout prediction module and the layout-conditioned generation module within a single unified multimodel foundation model. We adopt a training strategy that explicitly enforces cycle consistency across the framework through supervised fine-tuning (SFT) warm-up and cycle-consistent reinforcement learning (RL). Notably, the layout input used in our pipeline contains both precise bounding-box annotations and fine-grained textual descriptions for each object instance.

\subsection{Supervised Fine-tuning Warm-Up}

We first perform cycle-consistent supervised fine-tuning as a \emph{warm-up} stage to enable the model to reason over its own generated structures. This stage establishes a stable bidirectional mapping between layouts and images, which is essential for subsequent reinforcement learning where supervision is unavailable and the model must operate on self-generated intermediate representations.

Specifically, given an input image $I$ and its corresponding layout annotation $L$, the understanding objective minimizes the token-level cross-entropy between the predicted and ground-truth layout tokens.

To close the cycle, instead of conditioning generation on the ground-truth layout\cite{he2025plangen}, we explicitly feed the \emph{predicted} layout $\hat L$ from the understanding branch into the generation module. 

The generation objective minimizes the token-level cross-entropy between the predicted and ground-truth image tokens.

The SFT objective is defined as the sum of the understanding loss and the generation loss,
i.e., $L_{\mathrm{SFT}} = L_{\mathrm{Und}} + L_{\mathrm{Gen}}$,
ensuring both understanding and generation modules share a consistent representation space before RL. 

\begin{figure*}[t]
\centering

\includegraphics[width=0.75\linewidth]{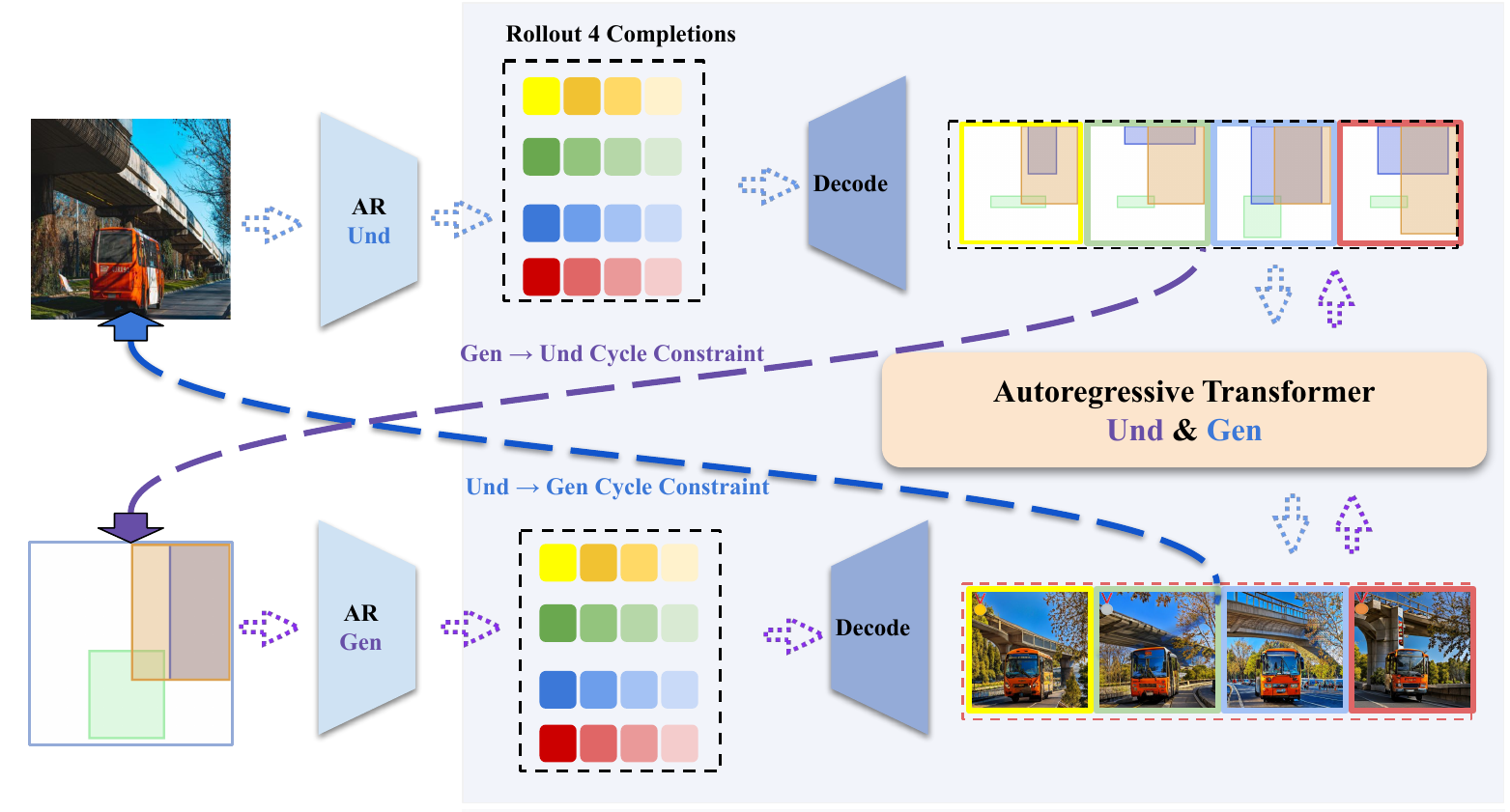}

\caption{
\textbf{Overview of CycleGRPO.} CycleGRPO trains a unified autoregressive transformer for both layout \textit{understanding} and layout-conditioned \textit{generation} through two complementary directions. The \textbf{Und $\rightarrow$ Gen} direction (\textcolor{blue}{blue arrows}) starts from an input image, where the model autoregressively predicts layouts via trajectory rollouts and decodes them into bounding boxes. The \textbf{Gen $\rightarrow$ Und} direction (\textcolor{violetarrow}{purple arrows}) starts from a layout prompt, where the model generates images and enforces that the generated images can recover the original layout through understanding. Multiple trajectories are sampled in both directions and optimized with GRPO, providing cycle-consistent reinforcement signals that align visual understanding with image generation.
}
\label{fig:overview}
\end{figure*}

\subsection{Cycle-consistent Reinforcement Learning}
Different from prior RL approaches that optimize either visual generation or visual understanding in isolation, we introduce \textbf{CycleGRPO}, a bidirectional cycle-consistent GRPO framework for interleaved unified multimodal modeling. 
Our formulation jointly optimizes the two complementary reasoning cycle: layout$\rightarrow$image$\rightarrow$layout generation and image$\rightarrow$layout$\rightarrow$image.
By interleaving these two cycle directions during training, the model learns not only to produce high-quality images or accurate layouts, but also to ensure that the generated visual structures remain semantically recoverable. 
This cycle constraint provides a form of \emph{self-introspection} between generation and understanding, which stabilizes reinforcement learning in unified multimodal models.

\paragraph{GRPO formulation}
DeepSeek R1-Zero \cite{guo2025deepseek} replaces supervised fine-tuning with reinforcement learning through its Group Relative Policy Optimization (GRPO) algorithm. 
Unlike PPO \cite{PPO}, which requires a separate critic network, GRPO estimates the policy advantage by comparing a group of sampled outputs without learning a value function.

Given an input $x$, the current policy $\pi_{\theta_{\mathrm{old}}}$ samples a group of $G$ candidate outputs
\[
\{o_1, \dots, o_G\} \sim \pi_{\theta_{\mathrm{old}}}(\cdot \mid x),
\]
each receiving a scalar reward $\{r_1,\dots,r_G\}$ from the reward function.
GRPO computes a relative advantage within the group by normalizing rewards with the empirical mean and standard deviation:
\begin{align}
A_i = \frac{r_i - \mathrm{mean}(\{r_j\}_{j=1}^G)}
{\mathrm{std}(\{r_j\}_{j=1}^G)} .
\end{align}

The policy is then optimized to increase the likelihood of responses with higher relative advantage. 
Let $y_i$ denote the token sequence of output $o_i$. 
The GRPO objective is defined as
\begin{align}
\mathcal{L}
=
-\mathbb{E}_{i}
\Big[
\min \big(
\rho_i A_i,\;
\mathrm{clip}(\rho_i, 1-\epsilon, 1+\epsilon) A_i
\big)
\Big]
+ \beta \, \mathrm{KL}(\pi_\theta \| \pi_{\theta_{\mathrm{ref}}}),
\end{align}
where
\[
\rho_i = 
\frac{\pi_\theta(y_i|x)}
{\pi_{\theta_{\mathrm{old}}}(y_i|x)}
\]
is the policy ratio between the updated policy and the behavior policy, 
$\epsilon$ controls the clipping range, and the KL penalty stabilizes optimization by constraining deviation from the reference policy.

Importantly, because GRPO operates solely on sampled outputs and scalar rewards—without requiring task-specific critics—it applies naturally to both directions of our unified model:  
\textbf{layout$\rightarrow$image generation} (optimizing visual quality and structural consistency) and \textbf{image$\rightarrow$layout understanding} (optimizing layout accuracy and interpretability).  

This symmetry enables an \textbf{interleaved RL optimization}, where rollout occurs alternately in the understanding and generation pathways under complementary reward signals. In practice, we use separate optimizer groups for different direction, denoted as $\theta_{\mathrm{Und}}$ and $\theta_{\mathrm{Gen}}$, which share the unified AR backbone and differ only in their generation and understanding encoder/decoder, without introducing any model-level modularity.

\paragraph{Understanding $\rightarrow$ Generation.}
In this
direction, the model first performs layout prediction, where multiple layout candidates are generated via rollout. The predicted layouts are then fed into the layout-to-image generation module, which proceeds to layout-to-image generation. The reward function evaluates both the structural accuracy of the predicted layouts and the perceptual quality of the generated images:
\begin{equation}
R_1 = \lambda_{\mathrm{iou}} \cdot \mathrm{IoU}(L_{\mathrm{pred}}, L_{\mathrm{gt}}) + 
\lambda_{\mathrm{clip}} \cdot \mathrm{CLIP}(I_{\mathrm{pred}}, I_{\mathrm{gt}}),
\end{equation}
where $\mathrm{IoU}$ measures geometric consistency between predicted and ground-truth layouts, and $\mathrm{CLIP}$ quantifies semantic consistence between generated and reference images. 

During this process, we optimize the $\theta_{\mathrm{Und}}$:
\begin{equation}
\theta_{\mathrm{Und}} \leftarrow \theta_{\mathrm{Und}} - \eta \nabla_{\theta_{\mathrm{Und}}}\mathcal{L}_{\mathrm{GRPO}}^{(1)},
\qquad
\end{equation}

\paragraph{ Generation $\rightarrow$ Understanding.}
In this direction, the model first performs layout-to-image generation by rolling out multiple image candidates conditioned on the input layout. The generated images are then passed to the understanding module to predict layouts. This reverse cycle encourages the generation module to synthesize images whose structural content can be reliably recovered. The reward therefore emphasizes both high-fidelity image synthesis and semantic recoverability:

\begin{equation}
R_2 = \lambda_{\mathrm{iou}} \cdot \mathrm{IoU}(L_{\mathrm{pred}}, L_{\mathrm{gt}}) +
\lambda_{\mathrm{hps}} \cdot \mathrm{HPSv2}(I_{\mathrm{pred}}),
\end{equation}
where $\mathrm{IoU}$ is similarly regulate the layout position and $\mathrm{HPSv2}$ is a high-level perceptual similarity metric capturing human-aligned aesthetic and semantic quality.

Similarly, we optimize the  $\theta_{\mathrm{Gen}}$: 
\begin{equation}
\theta_{\mathrm{Gen}} \leftarrow \theta_{\mathrm{Gen}} - \eta \nabla_{\theta_{\mathrm{Gen}}}\mathcal{L}_{\mathrm{GRPO}}^{(2)},
\qquad
\end{equation}

\paragraph{Cycle-consistent GRPO Objective.}
For each direction, the GRPO loss optimizes the model toward higher reward preference via gradient regularized updates. R denotes the reward assigned to the generated output, while b is a baseline value:
\begin{equation}
\mathcal{L}_{\mathrm{GRPO}} = - \mathbb{E}_{(x,y)} \big[ \log P_{\theta}(y \mid x) \cdot (R - b) \big]]
\end{equation}

\section{Experiment}

\begin{table*}[!htp]
\centering
\caption{\small 
Comparison of diffusion-based and autoregressive {\bf layout-to-image generation} models on the {\bf OverlayBench}~\cite{li2025overlaybench} benchmark across Simple, Regular, and Complex splits. Improvements are computed relative to PlanGen.
}

\label{tab:final_plangen_ours}
\resizebox{\linewidth}{!}
{
\setlength{\tabcolsep}{1pt}
\renewcommand{\arraystretch}{1.1}
\begin{tabular}{@{}clcccccccc@{}}
\toprule
\textbf{Type} &\textbf{Method} & \textbf{\# Params} 
& mIoU(\%) $\uparrow$ & O-mIoU(\%) $\uparrow$ 
& SR$_\text{E}$(\%) $\uparrow$ & SR$_\text{R}$(\%) $\uparrow$ 
& CLIP$_\text{Global}$ $\uparrow$ & CLIP$_\text{Local}$ $\uparrow$ \\
\midrule


\rowcolor[RGB]{245,245,245} 
\multicolumn{10}{c}{\textbf{OverlayBench-Simple}} \\

\multirow{5}{*}{Diffusion} 
& GLIGEN~\cite{li2023gligen}                & 1.07B  
& 60.54 & 36.22 & 49.99 & 78.72 & 34.17 & 24.75 \\

& InstanceDiff~\cite{wang2024instancediffusion}          & 1.23B  
& \textbf{71.21} & \textbf{49.99} & 77.71 & 87.49 & 34.25 & 27.69 \\

& MIGC~\cite{zhou2024migc}                  & 0.86B  
& 58.64 & 32.15 & 63.41 & 81.60 & 33.07 & 26.49 \\

& HiCo~\cite{cheng2024hico}                  & 1.22B  
& 69.47 & 47.23 & 67.75 & 86.08 & 35.25 & 27.04 \\

& CreatiLayout~\cite{zhang2024creatilayout} & 3.31B 
& 58.78 & 32.52 & 72.34 & 84.45 & \textbf{37.29} & 27.49 \\

\hdashline

\multirow{3}{*}{AR} 
& PlanGen~\cite{he2025plangen}  & 1.5B 
& 62.94 & 38.54 & 82.62 & 90.53 & 35.62 & 27.73 \\

& RL w/o Cycle & 1.5B 
& 57.02 \textcolor{red}{\scriptsize (-5.92)}
& 31.26 \textcolor{red}{\scriptsize (-7.28)}
& 71.38 \textcolor{red}{\scriptsize (-11.24)}
& 89.82 \textcolor{red}{\scriptsize (-0.71)}
& 34.05 \textcolor{red}{\scriptsize (-1.57)}
& 25.65 \textcolor{red}{\scriptsize (-2.08)} \\

& \textbf{\ours (ours)} & 1.5B 
& 64.84 \textcolor{ForestGreen}{\scriptsize (+1.90)}
& 42.03 \textcolor{ForestGreen}{\scriptsize (+3.49)}
& \textbf{85.05} \textcolor{ForestGreen}{\scriptsize (+2.43)}
& \textbf{90.53} \textcolor{ForestGreen}{\scriptsize (+0.00)}
& 35.48 \textcolor{red}{\scriptsize (-0.14)}
& \textbf{27.89} \textcolor{ForestGreen}{\scriptsize (+0.16)} \\

\midrule


\rowcolor[RGB]{245,245,245} 
\multicolumn{10}{c}{\textbf{OverlayBench-Regular}} \\

\multirow{5}{*}{Diffusion}
& GLIGEN~\cite{li2023gligen}                & 1.07B  
& 52.46 & 26.53 & 44.88 & 77.46 & 33.93 & 23.42 \\

& InstanceDiff~\cite{wang2024instancediffusion}          & 1.23B  
& \textbf{60.08} & \textbf{34.15} & 72.51 & 83.36 & 33.09 & 26.19 \\

& MIGC~\cite{zhou2024migc}                  & 0.86B  
& 47.42 & 20.06 & 56.67 & 77.85 & 32.72 & 24.99 \\

& HiCo~\cite{cheng2024hico}                  & 1.22B  
& 55.02 & 29.60 & 58.24 & 79.89 & 33.91 & 25.34 \\

& CreatiLayout~\cite{zhang2024creatilayout} & 3.31B 
& 47.04 & 20.67 & 62.60 & 78.31 & \textbf{36.67} & 25.55 \\

\hdashline

\multirow{3}{*}{AR} 
& PlanGen~\cite{he2025plangen} & 1.5B 
& 53.24 & 27.57 & 78.02 & 87.40 & 35.03 & 26.25 \\

& RL w/o Cycle & 1.5B 
& 46.37 \textcolor{red}{\scriptsize (-6.87)}
& 21.02 \textcolor{red}{\scriptsize (-6.55)}
& 65.76 \textcolor{red}{\scriptsize (-12.26)}
& 86.12 \textcolor{red}{\scriptsize (-1.28)}
& 33.18 \textcolor{red}{\scriptsize (-1.85)}
& 24.46 \textcolor{red}{\scriptsize (-1.79)} \\

& \textbf{\ours (ours)} & 1.5B 
& 54.21 \textcolor{ForestGreen}{\scriptsize (+0.97)}
& 29.04 \textcolor{ForestGreen}{\scriptsize (+1.47)}
& \textbf{80.58} \textcolor{ForestGreen}{\scriptsize (+2.56)}
& \textbf{87.98 }\textcolor{ForestGreen}{\scriptsize (+0.58)}
& 34.74 \textcolor{red}{\scriptsize (-0.29)}
& \textbf{26.42} \textcolor{ForestGreen}{\scriptsize (+0.17)} \\

\midrule


\rowcolor[RGB]{245,245,245} 
\multicolumn{10}{c}{\textbf{OverlayBench-Complex}} \\

\multirow{5}{*}{Diffusion}
& GLIGEN~\cite{li2023gligen}                & 1.07B  
& 50.79 & 23.85 & 41.70 & 79.93 & 33.92 & 22.75 \\

& InstanceDiff~\cite{wang2024instancediffusion}          & 1.23B  
& \textbf{53.68} & \textbf{25.63} & 66.02 & 80.34 & 32.33 & 25.53 \\

& MIGC~\cite{zhou2024migc}                  & 0.86B  
& 40.04 & 13.26 & 47.80 & 74.48 & 31.93 & 24.20 \\

& HiCo~\cite{cheng2024hico}                  & 1.22B  
& 46.56 & 20.35 & 48.88 & 75.19 & 33.15 & 24.41 \\

& CreatiLayout~\cite{zhang2024creatilayout} & 3.31B 
& 44.24 & 18.05 & 52.10 & 79.98 & \textbf{36.55} & 24.76 \\

\hdashline

\multirow{3}{*}{AR} 
& PlanGen~\cite{he2025plangen} & 1.5B 
& 49.87 & 22.94 & 72.82 & 86.80 & 35.09 & 25.37 \\

& RL w/o Cycle & 1.5B 
& 43.69 \textcolor{red}{\scriptsize (-6.18)}
& 17.08 \textcolor{red}{\scriptsize (-5.86)}
& 56.71 \textcolor{red}{\scriptsize (-16.11)}
& 87.13 \textcolor{ForestGreen}{\scriptsize (+0.33)}
& 33.07 \textcolor{red}{\scriptsize (-2.02)}
& 23.53 \textcolor{red}{\scriptsize (-1.84)} \\

& \textbf{\ours (ours)} & 1.5B 
& 50.72 \textcolor{ForestGreen}{\scriptsize (+0.85)}
& 24.50 \textcolor{ForestGreen}{\scriptsize (+1.56)}
& \textbf{74.01} \textcolor{ForestGreen}{\scriptsize (+1.19)}
& \textbf{87.07} \textcolor{ForestGreen}{\scriptsize (+0.27)}
& 34.57 \textcolor{red}{\scriptsize (-0.52)}
& \textbf{25.58} \textcolor{ForestGreen}{\scriptsize (+0.21)} \\

\bottomrule
\end{tabular}
}
\end{table*}

\subsection{Experiment Setup}
\paragraph{Training}
We sample our training data on LayoutSAM~\cite{zhang2024creatilayout}, which provides not only precise bounding-box annotations but also rich, fine-grained textual descriptions for each object instance. This combination of structured geometry and detailed semantics offers a more informative supervision signal than layouts alone, enabling our model to jointly learn spatial reasoning and object-level visual grounding. Following PlanGen~\cite{he2025plangen}, all images are resized to $384\times384$,  and layouts are tokenized into sequence form using our vision–language tokenizer.

\paragraph{Implementation details.}
All experiments are conducted on 8$\times$A6000 GPUs. 
We adopt mixed-precision training with per-GPU batch size 1 and gradient accumulation of 1 steps. 
The SFT phase runs for 7,200 steps with a learning rate of $2\!\times\!10^{-5}$, followed by 2,000 steps of GRPO training. For GRPO, we set the rollout size to 4, with a per-GPU batch size of 1 and 2 gradient accumulation steps. For rewards, we set $\lambda_{\text{iou}}\!=\!1$, $\lambda_{\text{clip}}\!=\!1$ and $\lambda_{\text{hps}}\!=2\!$.

\subsubsection{Evaluation}

We report metrics for both structural fidelity and perceptual quality:
We quantitatively evaluate our model on all three key capabilities: 
\textit{image captioning}, \textit{layout understanding} (Und), and \textit{layout-to-image generation} (Gen). 
We report results at the end of the SFT stage as well as after completing the CycleGRPO training. 
All evaluations are conducted on the held-out test splits of LayoutSAM and OverlayBench, 
demonstrating consistent improvements across caption quality, structural understanding accuracy, 
and layout-conditioned generation fidelity.

\subsection{Quantitative and Qualitative Results on Image Generation}

\subsubsection{OverlayBench}

\begin{figure*}[!h]
    \centering
    
    \includegraphics[width=\linewidth]{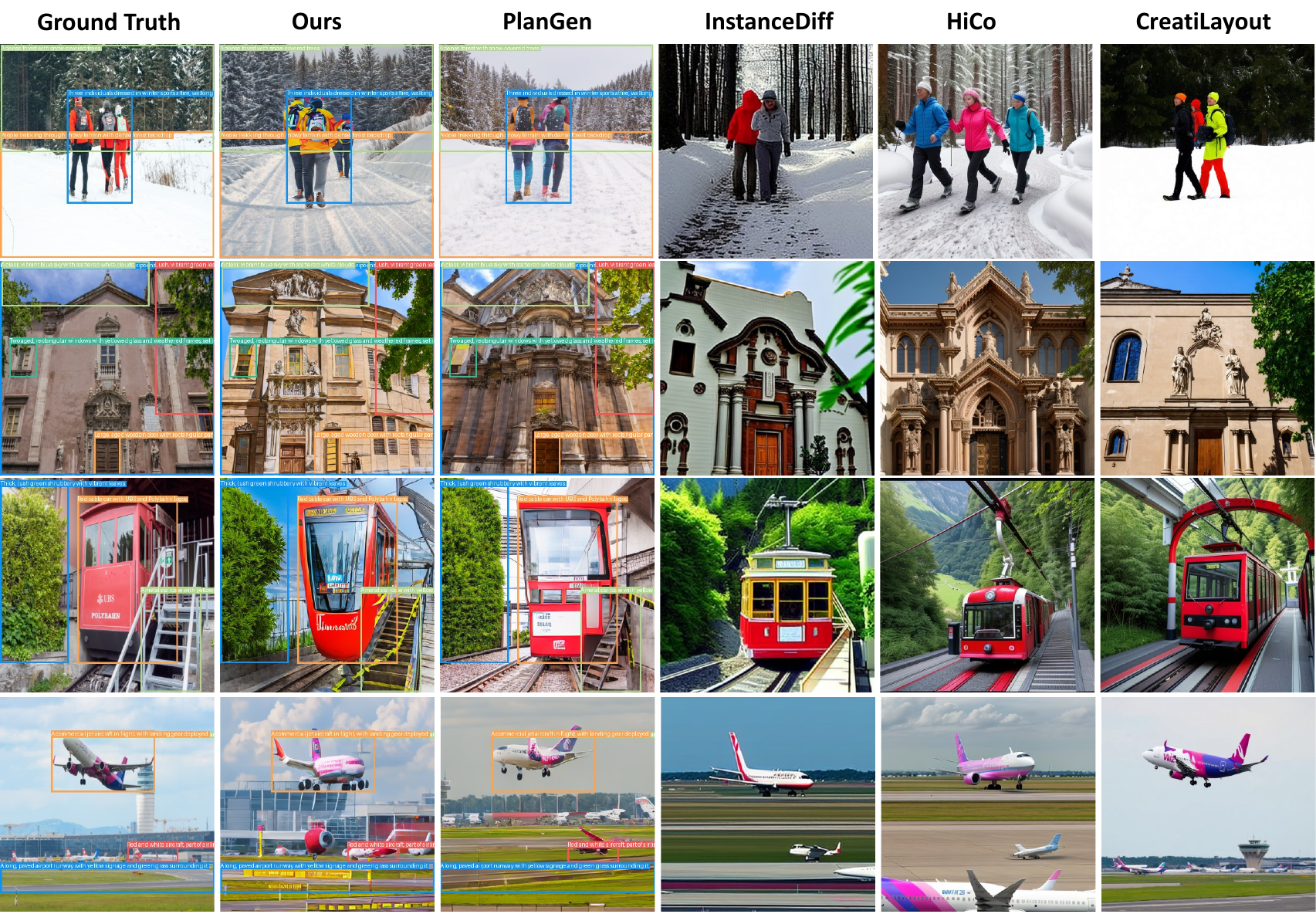} 
    \caption{\small \textbf{Qualitative comparison of layout-to-image generation.} 
Given the same input layouts (Ground Truth), our \ours model produces images that more faithfully 
preserve object geometry, spatial configuration, and fine-grained structure compared to PlanGen and 
diffusion-based baselines (InstanceDiff\cite{wang2024instancediffusion}, HiCo\cite{cheng2024hico}, CreatiLayout\cite{zhang2024creatilayout}). Across diverse scenes—including people, 
architectural facades, cable cars, and airplanes—\ours demonstrates stronger adherence to layout 
constraints and generates visually coherent, detail-rich images.}

    \label{fig:gen}
\end{figure*}

To comprehensively assess the robustness of layout-to-image (L2I) generation models under challenging real-world conditions, we conduct detailed evaluations on the OverlayBench~\cite{li2025overlaybench} benchmark, which provides three difficulty splits—Simple, Regular, and Complex—derived from the OverLayScore that quantifies spatial clutter, object density, and semantic ambiguity. While prior benchmarks such as COCO\cite{coco} and LayoutSAM\cite{zhang2024creatilayout} largely feature well-separated objects and low-overlap layouts, their limited structural diversity makes it difficult to stress-test the spatial reasoning and fine-grained grounding capabilities of modern L2I systems. In contrast, OverlayBench is specifically curated to expose systematic weaknesses: its harder splits contain dense multi-object compositions with substantial bounding box overlap, frequent inter-object occlusion, and high semantic similarity between co-occurring instances. Overlaybench introduces several challenging indicators. For instance, O-mIoU (Overlap mIoU) computes the mIoU restricted to ground-truth overlap regions and their corresponding predicted regions. By isolating areas shared between instances, O-mIoU provides a more sensitive and discriminative measure of fidelity in occluded or entangled scenarios than standard global mIoU.
SRR (Success Rate of Relationship) measures the percentage of object pairs whose predicted spatial relationships match the ground truth, offering an interpretable, relationship-level metric of success. SRE (Success Rate of Entity) measures the percentage of entities whose predicted attributes (e.g., category, location, or existence) correctly match the ground truth. These characteristics create severe challenges for both diffusion-based methods (e.g., GLIGEN~\cite{li2023gligen}, InstanceDiff~\cite{wang2024instancediffusion}, MIGC~\cite{zhou2024migc}, HiCo~\cite{cheng2024hico}, CreatiLayout~\cite{zhang2024creatilayout}) and autoregressive approaches such as PlanGen~\cite{he2025plangen}. As shown in \cref{tab:final_plangen_ours}, our supervised fine-tuning already improves spatial alignment and object-level consistency over the PlanGen baseline across all difficulty levels. Moreover, our CycleGRPO reinforcement learning strategy further enhances structure preservation and relation fidelity, particularly on the Regular and Complex splits where spatial entanglement is most pronounced. These results demonstrate that the proposed cycle-consistent training pipeline not only mitigates common failure modes under severe layout complexity, but also closes the performance gap between autoregressive and strong diffusion-based models in highly entangled scenes. The original image resolution of our output is 384. For fair comparison, we resize the image resolution to 512, the same as the baselines.

Notably, we further investigate the effect of removing the cycle-consistency constraint during reinforcement learning.
Without the bidirectional cycle supervision between layout understanding and layout-conditioned image generation, the RL optimization tends to focus on improving individual objectives while neglecting the structural alignment between the two tasks. As a result, the model exhibits noticeable performance degradation across multiple metrics compared with the PlanGen baseline.
This observation highlights that naive reinforcement learning without the cycle constraint can introduce optimization instability and task interference, ultimately harming the model's overall performance.
In contrast, our cycle-consistent training explicitly enforces mutual consistency between understanding and generation, providing a stabilizing signal that leads to consistent improvements.

\begin{table*}[!htp]
\centering
\caption{
{\bf Layout-to-image} quantitative comparison on the {\bf LayoutSAM-Eval \cite{zhang2024creatilayout}} benchmark. 
We report region-wise quality (Spatial, Color, Texture, Shape) and global perceptual metrics (PickScore, FID, IS). 
}
\setlength{\tabcolsep}{1em}
\resizebox{\linewidth}{!}{
\begin{tabular}{@{}clccccccc@{}}
\toprule
\multirow{2}{*}{Type} & \multirow{2}{*}{Method} 
& \multicolumn{4}{c}{Region-wise Quality} 
& \multicolumn{3}{c}{Global-wise Quality} \\ 
\cmidrule(lr){3-6} \cmidrule(l){7-9}
& & Spatial~$\uparrow$ & Color~$\uparrow$ & Texture~$\uparrow$ & Shape~$\uparrow$  
& Pick~$\uparrow$ & FID~$\downarrow$ & IS~$\uparrow$ \\ 
\midrule

& \gray{Real Images}          
& \gray{98.95} & \gray{98.45} & \gray{98.90} & \gray{98.80} 
& - & - & - \\  
\midrule

\multirow{5}{*}{Diffusion} 
& GLIGEN~\cite{li2023gligen}             
& 77.53 & 49.41 & 55.29 & 52.72  
& 20.78 & 21.92 & 20.57 \\

& InstanceDiff~\cite{wang2024instancediffusion}            
& 87.99 & 69.16 & 72.78 & 71.08  
& 21.01 & 19.67 & 20.02 \\

& MIGC~\cite{zhou2024migc}                
& 85.66 & 66.97 & 71.24 & 69.06  
& 20.71 & 21.19 & 19.65 \\

& Hico~\cite{cheng2024hico}         
& 90.22 & 69.89 & 73.95 & 72.75  
& \textbf{22.44} & 19.30 & \textbf{22.50} \\

& CreatiLayout~\cite{zhang2024creatilayout}         
& \underline{92.67} & 74.45 & 77.21 & 75.93  
& \underline{22.02} & \underline{19.10} & \underline{22.04} \\
\midrule

\multirow{2}{*}{AR}  
& PlanGen~\cite{he2025plangen}         
& 92.21 & \underline{82.69} & \underline{86.53} & \underline{85.36}  
& 21.43 & \textbf{13.91} & 19.31 \\

& \textbf{\ours (ours)}        
& \textbf{92.71} & \textbf{84.22} & \textbf{87.97} & \textbf{87.02}  
& 21.53 & \underline{15.62} & 20.42 \\

\bottomrule
\end{tabular}
}
\label{tab:g2i}
\end{table*}
\subsubsection{LayoutSAM-eval}
To evaluate fine-grained controllability and structural fidelity in layout-to-image generation, 
we benchmark diffusion-based and autoregressive models on the LayoutSAM-Eval dataset. 
Beyond standard global perceptual metrics such as CLIP, FID, and IS, this benchmark measures 
\emph{region-wise} consistency across Spatial, Color, Texture, and Shape attributes, providing 
a more sensitive diagnostic of how faithfully a model adheres to object-level layout constraints.

Diffusion models such as GLIGEN~\cite{li2023gligen}, InstanceDiff~\cite{wang2024instancediffusion}, MIGC~\cite{zhou2024migc}, HiCo~\cite{cheng2024hico}, CreatiLayout~\cite{zhang2024creatilayout} demonstrate strong overall 
image quality but still exhibit noticeable inconsistencies in regional appearance, especially under 
dense layouts with fine object boundaries. In contrast, autoregressive models inherently maintain 
token-level structural coupling, enabling more precise alignment between layout tokens and generated 
visual attributes. Our method further strengthens this alignment: despite using a unified AR framework, 
it achieves the highest scores across all four regional metrics, surpassing both diffusion-based 
baselines and the autoregressive PlanGen\cite{he2025plangen}. These results highlight the effectiveness of our cycle-consistent 
training and RL-enhanced layout grounding in preserving detailed object-level semantics. 
Global-level metrics for our model will be updated once the remaining evaluations are completed. We further present qualitative layout-to-image generation results in \cref{fig:gen}, 
which compares our method against PlanGen, InstanceDiff, HiCo, and CreatiLayout. 
Across diverse scenes, \ours produces images that more faithfully adhere to the input layouts, 
preserving object geometry, spatial relations, and appearance details. This result highlight the effectiveness of our framework.

\subsection{Quantitative and Qualitative Results on Image Understanding}

\subsubsection{Layout Understanding}

\begin{table}[!htp]
\centering
\caption{\small Quantitative results on \textbf{layout understanding} evaluated on the LayoutSamEval benchmark\cite{zhang2024creatilayout}. We compare our model with the PlanGen baseline, along with strong vision-language and detection baselines, including Grounding-DINO~\cite{liu2024grounding} and Janus-Pro\cite{chen2025janus}. Metrics include AP, AP50, AP75, and AR. Improvements are computed relative to PlanGen.}
\label{tab:und_result}
\setlength{\tabcolsep}{1.8em}
\resizebox{\linewidth}{!}{
\begin{tabular}{@{} lcccc @{}}
\toprule
\textbf{Setting} & \textbf{AP}↑ & \textbf{AP50}↑ & \textbf{AP75}↑ & \textbf{AR}↑\\
\midrule
\gray{GroundingDINO~\cite{liu2024grounding}} & \gray{42.53} & \gray{47.20} & \gray{44.60} & \gray{83.61} \\
Janus-Pro~\cite{chen2025janus} & 9.05 & 13.68 & 7.20 & 14.67\\
PlanGen-baseline~\cite{he2025plangen}
& 27.66 & 33.94 & 30.56 & 51.58 \\

\textbf{\ours (ours)}
& 28.33 \textcolor{ForestGreen}{\scriptsize ($+0.67$)}
& 36.99 \textcolor{ForestGreen}{\scriptsize ($+3.05$)}
& 30.81 \textcolor{ForestGreen}{\scriptsize ($+0.25$)}
& 71.05 \textcolor{ForestGreen}{\scriptsize ($+19.47$)} \\
\bottomrule
\end{tabular}
}
\end{table}

\begin{figure*}[!h]
    \centering

    \includegraphics[width=\linewidth]{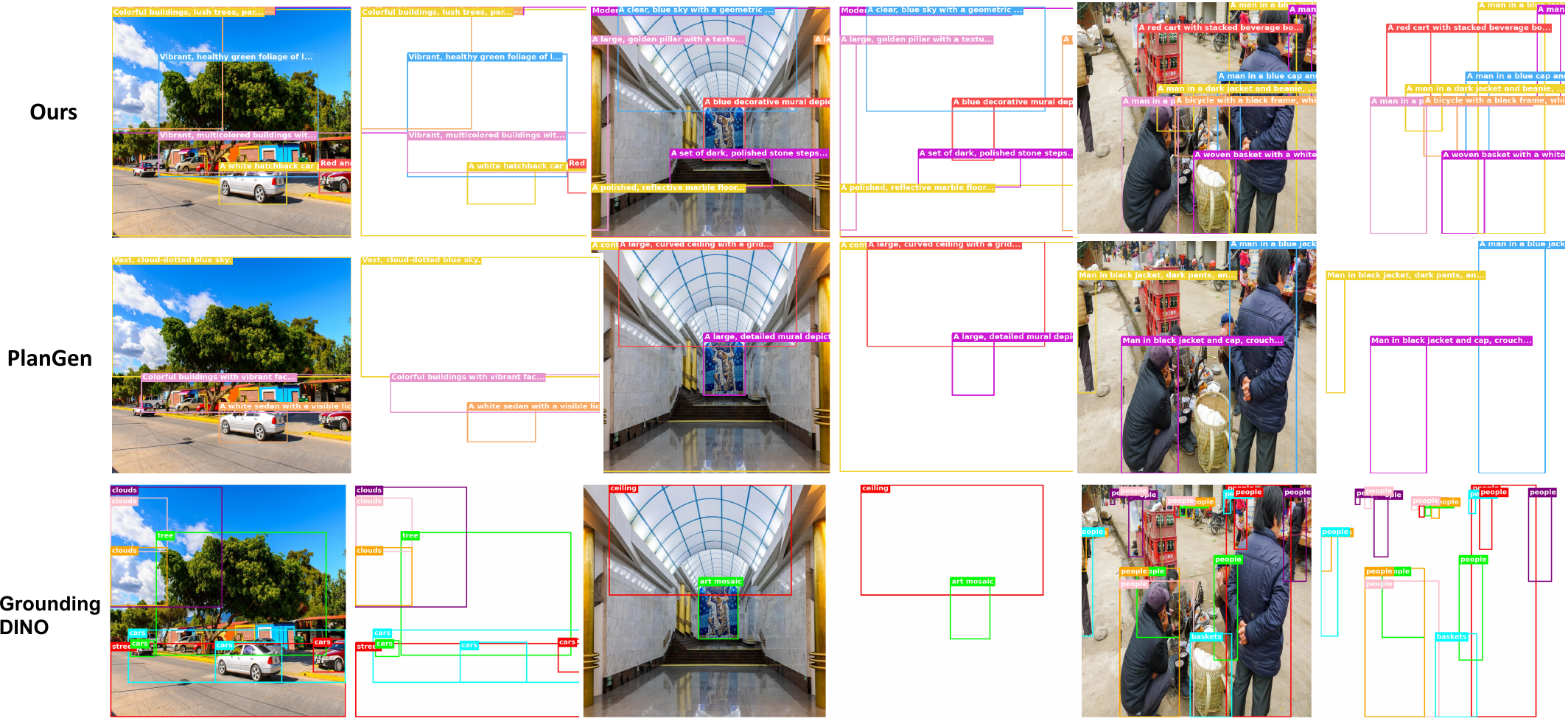} 
    \caption{\small \textbf{Qualitative comparison of layout understanding results.} 
    We compare \ours (top) with PlanGen (middle) and GroundingDINO (bottom) across diverse scenes. 
    \ours produces more complete and precise detections, with fewer missed objects, 
    cleaner bounding boxes, and more semantically coherent structured descriptions. 
    In contrast, PlanGen often yields fragmented boxes, duplicated predictions, and under-detection in complex scenes. 
    These results demonstrate that our cycle-consistent training substantially boosts structural grounding quality.  }
    \label{fig:detection_compare}
    \label{fig:und}
\end{figure*}

Following Plangen\cite{he2025plangen}, to assess the effectiveness of our approach in layout understanding, we evaluate the model on the LayoutSamEval benchmark and compare it with both autoregressive and non-autoregressive baselines. In addition to the original PlanGen layout parser, we incorporate Grounding~DINO and Janus Pro as strong detection-oriented and vision-language-oriented baselines. Grounding~DINO offers a high-quality object localization reference point, while Janus Pro represents a competitive multimodal large language model with robust grounding capabilities. As shown in \cref{tab:und_result}, \ours yields higher AP and dramatically improved AR, demonstrating stronger grounding completeness and more reliable extraction of layout structures. While AP emphasizes how accurate and well-ranked the predictions are, AR emphasizes whether the model finds all relevant objects at all. AR is therefore particularly informative in proposal-based, layout, and grounding-oriented detection settings, where complete spatial coverage is more critical than confidence ordering. These results confirm that our progressive training strategy not only enhances layout understanding within the autoregressive modeling framework but also brings its performance better. As illustrated in \cref{fig:und}, our model generates 
significantly more accurate and complete layout predictions than the PlanGen baseline. 
This indicates that our proposed cycle-consistent training effectively reduces missed detections, stabilizes bounding-box localization, 
and produces semantically coherent descriptions even in crowded or highly structured scenes.

\begin{table}[!h]
\centering
\caption{\small Quantitative comparison of the {\bf image captioning} performance on the LayoutSamEval benchmark\cite{zhang2024creatilayout}.
We report BLEU-1/2/3/4, METEOR, ROUGE-L, and CIDEr scores for the baseline PlanGen~\cite{he2025plangen} and Janus Pro~\cite{chen2025janus}, our SFT-variant model, and reinforcement learning. Note that \ours has {\bf not been trained on} this captioning task, yet, it demonstrates a substantial improvement over the baselines, clearly demonstrating the strong advantage of adding the cycle-consistency to the basic visual foundation model. Improvements are computed relative to PlanGen.}

\label{tab:caption}
\resizebox{\linewidth}{!}{
\setlength{\tabcolsep}{6pt}

\begin{tabular}{@{}lccccccc@{}}
\toprule
\textbf{Method} 
& \textbf{BLEU-1}↑
& \textbf{BLEU-2}↑
& \textbf{BLEU-3}↑
& \textbf{BLEU-4}↑ 
& \textbf{METEOR}↑ 
& \textbf{ROUGE-L}↑ 
& \textbf{CIDEr}↑ \\
\midrule

Janus-Pro~\cite{chen2025janus}
& 35.26 & 19.73 & 11.19 & 6.66 & 21.78 & 26.04 & 12.80 \\

PlanGen-baseline~\cite{he2025plangen}
& 43.06 
& 30.20 
& 22.68 
& 17.92 
& 21.92 
& 33.81 
& 37.48 \\

\textbf{\ours (ours)} {\scriptsize SFT}   
& 51.72 \textcolor{ForestGreen}{\scriptsize ($+8.66$)}
& 37.97 \textcolor{ForestGreen}{\scriptsize ($+7.77$)}
& 29.27 \textcolor{ForestGreen}{\scriptsize ($+6.60$)}
& 23.58 \textcolor{ForestGreen}{\scriptsize ($+5.66$)}
& 25.43 \textcolor{ForestGreen}{\scriptsize ($+3.51$)}
& 40.10 \textcolor{ForestGreen}{\scriptsize ($+6.29$)}
& 57.90 \textcolor{ForestGreen}{\scriptsize ($+20.42$)} \\

RL w/o cycle   
& 46.63 \textcolor{ForestGreen}{\scriptsize (+3.57)}
& 31.83 \textcolor{ForestGreen}{\scriptsize (+1.63)}
& 23.20 \textcolor{ForestGreen}{\scriptsize (+0.52)}
& 17.99 \textcolor{ForestGreen}{\scriptsize (+0.07)}
& 21.85 \textcolor{red}{\scriptsize (-0.07)}
& 34.92 \textcolor{ForestGreen}{\scriptsize (+1.11)}
& 38.86 \textcolor{ForestGreen}{\scriptsize (+1.38)} \\

\textbf{\ours (ours)} {\scriptsize RL}   
& 52.99 \textcolor{ForestGreen}{\scriptsize ($+9.93$)} 
& 39.91 \textcolor{ForestGreen}{\scriptsize ($+9.71$)} 
& 31.41 \textcolor{ForestGreen}{\scriptsize ($+8.73$)} 
& 25.68 \textcolor{ForestGreen}{\scriptsize ($+7.76$)} 
& 26.84 \textcolor{ForestGreen}{\scriptsize ($+4.92$)} 
& 42.35 \textcolor{ForestGreen}{\scriptsize ($+8.54$)} 
& 66.46 \textcolor{ForestGreen}{\scriptsize ($+28.98$)} \\

\bottomrule
\end{tabular}
}

\end{table}

\subsubsection{Image Captioning as An Emergent Task}
To demonstrate the {\bf cross-task generalization} capability of our \ours model, we show experimental results on an emergent task, image captioning, that is {\bf not included} in the reinforcement learning optimization.
As shown in \cref{tab:caption}, on the LayoutSamEval benchmark, we observe clear and consistent improvements across all captioning metrics as the model transitions from the original PlanGen baseline to our supervised fine-tuning (SFT) variant and subsequently to the CycleGRPO reinforcement learning procedure. This demonstrates the significance of \ours for its cross-task generalization capability with the universal benefit of introducing the cycle-consistency loss. The ablation study in \cref{tab:caption} further illustrates this. BLEU-1/2/3/4 measure n-gram precision between generated captions and references, with higher-order BLEU scores (e.g., BLEU-4) emphasizing longer and more fluent phrase matching. METEOR considers both precision and recall of word matches, while also accounting for stemming and synonymy, making it more sensitive to semantic similarity. ROUGE-L evaluates the longest common subsequence between generated and reference captions, capturing sentence-level fluency and structural consistency. CIDEr weights n-grams by their inverse document frequency and measures consensus with multiple references, emphasizing human-like and informative descriptions. While PlanGen provides a strong starting point, SFT substantially enhances lexical and structural alignment, reflected by notable gains in BLEU-1/2/3/4 and ROUGE-L. Building on this, our cycle-consistent reinforcement learning further improves semantic fidelity and descriptive richness, leading to higher METEOR and CIDEr scores and achieving the best overall performance across all metrics. These results collectively verify that simultaneously improving both the image understanding and the image generation capabilities in integration can indeed bring {\bf general and universal benefits to each other}.

To further disentangle the source of performance improvements, we conduct an ablation study on the image captioning task. As shown in Table~\ref{tab:caption}, removing the cycle constraint (``RL w/o cycle''), where reinforcement learning is applied only with the IoU reward for layout understanding, yields only marginal gains over the PlanGen baseline. For instance, BLEU-1 improves slightly from 43.06 to 46.63. More importantly, this variant consistently underperforms the SFT model across all metrics (e.g., CIDEr drops from 57.90 to 38.86 and ROUGE-L from 40.10 to 34.92), indicating that naive reinforcement learning without structural constraints can even harm the cross-task generalization ability of the model. 
In contrast, introducing our cycle-consistent reinforcement learning substantially improves performance across all metrics, achieving 52.99 BLEU-1, 39.91 BLEU-2, and a significant gain in CIDEr to 66.46. These observations suggest that the improvement does not simply come from additional RL optimization, but rather from the reciprocal supervision enforced by the cycle-consistent framework, which aligns layout understanding and image generation and provides a more stable learning signal.

\begin{table}[h]
\centering
\footnotesize
\caption{Comparison with PlanGen and our variants on \textbf{LayoutSAM-Eval\cite{zhang2024creatilayout}}. Understanding metrics are on the left, and generation metrics on the right.}
\resizebox{\linewidth}{!}{
\setlength{\tabcolsep}{1pt}
\renewcommand{\arraystretch}{0.8}

\begin{tabular}{@{} lcccc|cccc @{}}
\toprule
\textbf{Model} 
& \textbf{AP}$\uparrow$ 
& \textbf{AP50}$\uparrow$ 
& \textbf{AP75}$\uparrow$ 
& \textbf{AR}$\uparrow$  
& \textbf{Spatial}$\uparrow$ 
& \textbf{Color}$\uparrow$ 
& \textbf{Texture}$\uparrow$ 
& \textbf{Shape}$\uparrow$ \\ 
\midrule

PlanGen
&27.66 & 33.94 & 30.56 & 51.58 & 92.21 & 82.69 & 86.53 & 85.36 \\

PlanGen-{\scriptsize Und}
& 28.00 
& 37.78 
& 29.40 
& 48.17 
& 88.16 \down{-4.05}
& 76.92 \down{-5.77}
& 80.71 \down{-5.82}
& 79.71 \down{-5.65} \\

PlanGen-{\scriptsize Gen}
& 8.41  \down{-19.25}
& 16.33 \down{-17.61}
& 7.27  \down{-23.29}
& 17.62 \down{-33.96}
& 91.41 
& 81.83 
& 85.84 
& 84.81  \\

PlanGen-{\scriptsize Joint}
& 24.28 \down{-3.38}
& 34.26 \up{0.32}
& 25.38 \down{-5.18}
& 45.54 \down{-6.04}
& 91.00 \down{-1.21}
& 81.75 \down{-0.94}
& 85.64 \down{-0.89}
& 84.28 \down{-1.08}  \\

PlanGen-{\scriptsize w/o cycle}
& 27.83 \up{0.17}
& 36.77 \up{2.83}
& 29.63 \down{-0.93}
& 51.61 \up{0.03}
& 91.72 \down{-0.49}
& 82.45 \down{-0.24}
& 86.78 \up{0.25}
& 86.19 \up{0.83} \\

\textbf{\ours (ours)}
& \textbf{28.33} \textcolor{ForestGreen}{\scriptsize ($+0.67$)}
& \textbf{36.99} \textcolor{ForestGreen}{\scriptsize ($+3.05$)}
& \textbf{30.81} \textcolor{ForestGreen}{\scriptsize ($+0.25$)}
& \textbf{71.05} \textcolor{ForestGreen}{\scriptsize ($+19.47$)} 
& \textbf{92.71} \up{0.50}
& \textbf{84.22} \up{1.53}
& \textbf{87.97} \up{1.44}
& \textbf{87.02} \up{1.66} \\

\bottomrule
\end{tabular}
}

\label{tab:plangen_ablation}
\end{table}

\subsection{Ablation}

Our ablation results further demonstrate the necessity of cycle-consistent modeling. 
\cref{tab:plangen_ablation} summarizes different training configurations compared with the PlanGen baselines: PlanGen-Und fine-tunes the model exclusively on the visual understanding objective, while PlanGen-Gen fine-tunes on the layout-to-image generation objective in a next token prediction paradiam. PlanGen-Joint applies interleaved joint training of understanding and generation just as Plangen baseline. PlanGen-w/o cycle adopts the same training strategy as us, which builds on SFT warm-up and RL training (IOU reward for Layout Understanding and HPSv2 for layout-to-image Generation)for understanding and generation without explicitly enforcing cycle consistency.  
The table first reveals a clear phenomenon of SFT forgetting: optimizing the model for visual understanding improves detection but degrades generation fidelity (PlanGen-Und), while optimizing for generation severely weakens visual understanding (PlanGen-Gen). This trade-off 
highlights the limitation of naive or one-sided fine-tuning strategies. Notably, although PlanGen-w/o cycle partially alleviates this issue by jointly applying RL on top of an SFT-initialized model, it still fails to consistently preserve bidirectional competence and even encount forgetting problem over baselins, indicating that apply reinforcement learning optimization alone is insufficient. In contrast, explicitly enforcing cycle consistency on the model’s own predictions (\ours), rather than relying on interleaved joint training (PlanGen-Joint) or solely reinforcement learning training(PlanGen-w/o cycle), leads to consistent and robust improvements over an already \textbf{converged} PlanGen baseline.

\section{Conclusion}
We presented \ours, a unified framework that introduces cycle-consistency to visual foundation models for image understanding and generation within an integrated autoregressive formulation. By introducing CycleGRPO reinforcement strategy, our model reduces the mutual interference that commonly appears in unified vision--language systems and achieves more stable bidirectional alignment. Experiments across layout prediction, controllable generation, and emergent captioning tasks show clear and consistent improvements over the PlanGen baseline, highlighting the effectiveness and generalization capabilities of enforcing reciprocal supervision between inverse tasks. We believe our cycle-consistent formulation offers a simple yet powerful direction for building unified multimodal models that can better align, self-correct, and generalize across understanding and generation tasks.

\input{sec/acknowledgment}
{
    \small
    \bibliographystyle{splncs04}
    \bibliography{main}
}
\newpage
\input{sec/supplementary}

\end{document}

%% file: preamble.tex
\usepackage[pagebackref]{hyperref}

\usepackage{eccvabbrv}
\usepackage{graphicx}
\usepackage{booktabs}
\usepackage{pifont}
\usepackage{multirow}
\usepackage{colortbl}
\usepackage{paralist}
\usepackage{mathtools}

\usepackage{arydshln}
\usepackage[accsupp]{axessibility}
\usepackage{orcidlink}

\newcommand{\gray}[1]{\textcolor{gray}{#1}}
\newcommand{\up}[1]{\textcolor{green!70!black}{\scriptsize(+#1)}}
\newcommand{\down}[1]{\textcolor{red!70!black}{\scriptsize(#1)}}

\footnotetext[2]{$^*$ Equal contribution.\\$^\dag$ Contributed to this work during an internship at UC San Diego.}
\def\ours{CyCLeGen\xspace}

\title{\ours:
Cycle-Consistent\texorpdfstring{\\}{} Layout Prediction and Image Generation\texorpdfstring{\\}{} in Vision Foundation Models}
\titlerunning{\ours}
\author{
Xiaojun Shan,\textsuperscript{*} 
Haoyu Shen,\textsuperscript{*\textdagger} 
Yucheng Mao, 
Xiang Zhang, 
Anand Abhay,
Bingnan Li,
Haiyang Xu,
Zhuowen Tu
}
\authorrunning{Shan et al.}
\institute{UC San Diego}

%% file: sec/acknowledgment.tex
\section*{Acknowledgment}
This work is supported by NSF award IIS-2127544 and NSF award IIS-2433768. We thank Divyansh Srivastava, Ethan Armand, and Zeyuan Chen for helpful discussions.

\clearpage

%% file: sec/supplementary.tex
\appendix
\section{Additional Result}

\subsection{Quantitative Result for Image Generation}

\begin{table}[h]
\centering
\caption{HPSv2 comparison between diffusion-based methods and autoregressive models.}
\label{tab:hpsv2}
\small
\renewcommand{\arraystretch}{1.1}
\setlength{\tabcolsep}{4pt}

\begin{tabular}{l l c}
\toprule
\textbf{Type} & \textbf{Method} & \textbf{HPSv2} \\
\midrule

\multirow{3}{*}{Diffusion}
& InstanceDiff~\cite{wang2024instancediffusion} & 21.25 \\
& HiCo~\cite{cheng2024hico} & 27.51 \\
& CreatiLayout~\cite{zhang2024creatilayout} & 25.95 \\

\midrule

\multirow{2}{*}{AR}
& PlanGen~\cite{he2025plangen} & 24.50 \\
& \textbf{\ours{} (ours)} & \textbf{25.96} \\

\bottomrule
\end{tabular}
\end{table}

To further enhance perceptual quality during image generation, our RL process integrates HPSv2~\cite{wu2023human} as a verifiable reward. HPSv2 is a human-aligned perceptual scoring metric designed to evaluate generated images based on their overall aesthetic fidelity, realism, and semantic consistency. Unlike pixel-level or embedding-level metrics such as FID or CLIP score, HPSv2 captures holistic human preference signals, making it particularly suitable for guiding the optimization of generative models. When used as a reward in our reinforcement learning, HPSv2 effectively encourages the model to produce images that not only respect the underlying layout structure but also exhibit stronger perceptual appeal. As shown in \cref{tab:hpsv2}, our method achieves an HPSv2 score of \textbf{25.96}, outperforming the PlanGen baseline (\textbf{24.50}) by a significant margin. This demonstrates that perceptual reinforcement with HPSv2 leads to consistently higher-quality outputs, validating the benefit of incorporating human-aligned perceptual rewards into autoregressive layout-to-image generation.

\subsection{Quantitative Result for Layout Planning and Image Generation}

\begin{table}[h]
\vspace{-7pt}
\centering

\begin{tabular}{@{}lccccc}
\toprule
Method & Pick$\uparrow$  & FID $\downarrow$ & IS$\uparrow$ \\
\midrule
\gray{GT Layout} &\gray{21.53} &\gray{15.62} &\gray{20.42} \\
\midrule

\midrule
PlanGen &21.40  &16.18 &20.18 \\
\ours{} &21.32  &19.27 &\textbf{21.05} \\

\bottomrule
\end{tabular}
\caption{
    Layout planning quantitative comparison of \ours{} with Plangen\cite{he2025plangen} and GT layout on LayoutSAM-Eval\cite{zhang2024creatilayout}.
}
\label{tab:plan}
\vspace{-12pt}
\end{table}
Beyond layout understanding and layout-to-image generation, we further evaluate whether our model can generalize to the layout planning task, despite not receiving any task-specific training or supervision for planning. As shown in \cref{tab:plan}, our method achieves a competitive PickScore and clearly improves the Inception Score (IS) over PlanGen. Notably, PlanGen is explicitly trained with layout planning data, whereas our model learns no planning objective at all—yet it still produces high-quality layouts that align well with the ground-truth distribution. This highlights the strong inductive bias introduced by our cycle-consistent training: once the model acquires robust bidirectional layout--image reasoning, it naturally inherits the ability to plan plausible layouts without dedicated supervision.
\subsection{Ablation on SFT}
\begin{table*}[h]
\centering
\setlength{\tabcolsep}{6pt}
\renewcommand{\arraystretch}{1.0}
\caption{Comparison between PlanGen, training without SFT, and \ours. Understanding metrics are on the left, and generation metrics on the right. Improvements are computed relative to the case without SFT.}

\resizebox{\linewidth}{!}{
\setlength{\tabcolsep}{1pt}
\begin{tabular}{lcccc|cccc}
\toprule
\textbf{Setting} 
& \textbf{AP}$\uparrow$ 
& \textbf{AP50}$\uparrow$ 
& \textbf{AP75}$\uparrow$ 
& \textbf{AR}$\uparrow$ 
& \textbf{Spatial}$\uparrow$ 
& \textbf{Color}$\uparrow$ 
& \textbf{Texture}$\uparrow$ 
& \textbf{Shape}$\uparrow$ \\
\midrule
PlanGen
& 27.66 & 33.94 & 30.56 & 51.58
& 92.21 & 82.69 & 86.53 & 85.36 \\

w / o SFT
& 27.89 & 36.77 & 30.62 & 62.91
& 92.69 & 83.79 & 87.29 & 86.46 \\

\textbf{\ours}
& \textbf{28.33} 
& \textbf{36.99} 
& \textbf{30.81} 
& \textbf{71.05} 
& \textbf{92.71} 
& \textbf{84.22} 
& \textbf{87.97} 
& \textbf{87.02} \\
\midrule

$\Delta$ 
& \textcolor{ForestGreen}{\textbf{+0.44}}
& \textcolor{ForestGreen}{\textbf{+0.22}}
& \textcolor{ForestGreen}{\textbf{+0.19}}
& \textcolor{ForestGreen}{\textbf{+8.14}}
& \textcolor{ForestGreen}{\textbf{+0.02}}
& \textcolor{ForestGreen}{\textbf{+0.43}}
& \textcolor{ForestGreen}{\textbf{+0.68}}
& \textcolor{ForestGreen}{\textbf{+0.56}} \\
\bottomrule
\end{tabular}
}
\label{tab:pg_undrl_genrl_comparison}
\end{table*}

As shown in \cref{tab:pg_undrl_genrl_comparison}, removing the supervised fine-tuning (SFT) warm-up and directly applying cycle-consistent reinforcement learning on top of PlanGen (w / o SFT) leads to relatively modest improvements compared to the full pipeline. While w / o SFT improves over PlanGen on several understanding metrics, the gains are smaller than those achieved by our method with SFT. In particular, incorporating SFT yields consistent improvements across all understanding metrics, with notable gains in AR, and further enhances generation quality in terms of spatial, color, texture, and shape consistency. These results suggest that SFT provides an effective initialization that facilitates subsequent cycle-consistent RL optimization, resulting in more stable and comprehensive performance improvements.

\subsection{Image Editing and Object Removal Evaluation}

\begin{table}[t]
\centering
\caption{Comparison of image editing performance between PlanGen and \ours{}. Higher is better for all metrics.}
\label{tab:edit}
\scalebox{0.9}{
\begin{tabular}{lccc}
\toprule
\textbf{Method} & \textbf{LocalCLIP} $\uparrow$ & \textbf{IS} $\uparrow$ & \textbf{SR}(\%)  $\uparrow$ \\
\midrule
PlanGen  & 25.28 & 9.08 & 14.50 \\
CycleGen & \textbf{25.31} & \textbf{9.18} & \textbf{17.00} \\
\bottomrule
\end{tabular}
}
\end{table}

\begin{table}[t]
\centering
\caption{Object removal results on 200 samples from the COCO test set.
Higher is better for all metrics, except for LocalCLIP.}
\label{tab:rm}
\scalebox{0.9}{
\begin{tabular}{lccc}
\toprule
\textbf{Method} & \textbf{LocalCLIP} $\downarrow$ & \textbf{IS} $\uparrow$ & \textbf{SR} (\%) $\uparrow$ \\
\midrule
PlanGen  & 23.08 & \textbf{9.37} & \textbf{87.50} \\
CycleGen & \textbf{22.92} & 9.09 & 86.00 \\
\bottomrule
\end{tabular}
}
\end{table}

We evaluate the image editing capability of our cycle-based framework by comparing \textbf{PlanGen} and \ours{} on two representative tasks: \emph{general image editing} and \emph{object removal}. All methods are evaluated under identical generation settings to ensure a fair comparison.

Following the evaluation protocol adopted in PlanGen, we report three complementary metrics: 
(1) \textbf{LocalCLIP}, which measures the semantic alignment between edited local regions and the corresponding textual instructions, reflecting locality-aware controllability; 
(2) \textbf{Inception Score (IS)}, which evaluates the overall visual quality and realism of the generated images; 
(3) \textbf{Success Rate (SR)}, which measures whether the intended editing objective is successfully achieved. 
In particular, SR is computed using MiniCPM as an \emph{LLM-as-a-Judge}, where the model is prompted to determine whether the desired editing effect (e.g., object removal or attribute modification) is correctly realized in the generated image.

\paragraph{Image Editing Results.}
As shown in \cref{tab:edit}, CycleGen consistently outperforms PlanGen across all metrics on the general image editing task.
Notably, CycleGen achieves a higher SR, indicating a more reliable execution of editing instructions.
Meanwhile, improvements in LocalCLIP and IS demonstrate that the cycle-consistent optimization not only enhances semantic alignment within edited regions but also preserves or slightly improves overall image quality.
These results suggest that enforcing a closed-loop consistency between layout planning and image generation effectively mitigates instruction drift and partial edit failures, which are common in open-ended image editing scenarios.

\paragraph{Object Removal Results.}
We further evaluate object deletion performance on 200 object removal cases derived from the COCO test set.
As reported in \cref{tab:rm}, both PlanGen and CycleGen achieve strong performance, with high SR values, indicating that both methods are effective at removing target objects.
PlanGen slightly outperforms CycleGen in terms of SR and IS, while CycleGen achieves comparable LocalCLIP scores.
This observation suggests that for well-constrained editing tasks such as object removal, where the target region and objective are explicit, the benefit of cycle-based optimization is less pronounced, though it does not degrade deletion fidelity or image realism.

Overall, these results demonstrate that CycleGen provides more stable and reliable performance for complex image editing tasks requiring precise semantic alignment, while maintaining competitive object removal capability and visual quality.

\section{Limitations}
Although our unified cycle-consistent framework demonstrates strong improvements in both layout understanding and layout-conditioned generation, two limitations remain.

\textbf{Model capacity constraints.}
Our current implementation is built on a 1.5B-parameter Plangen\cite{he2025plangen} backbone, which is significantly smaller than contemporary vision–language foundation models.  
While this lightweight design enables efficient training and experimentation, it also limits the model’s ability to perform 
VQA and open-ended reasoning tasks. 

\textbf{Quality of RL data.}
Our reinforcement learning stage relies on data from open-source dataset LayoutSAM\cite{zhang2024creatilayout}, which is labeled by Grounding-Dino\cite{liu2024grounding}, which may contain structural inaccuracies, visual artifacts, or alignment biases.  
Although our GRPO procedure improves robustness, the learning signal is ultimately constrained by the quality of these synthetic samples.  
Improving the fidelity of RL data, either through stronger generative models, human preference signals, or automated data filtering—represents an important direction for enhancing overall performance.

\textbf{Generation diversity}
Cycle consistency may be less effective in scenes with dense or ambiguous layouts, where strong structural constraints can limit generation diversity. 

Despite these limitations, our model is a universal framework that can be applied to different base models. Our results also highlight the promise of cycle-consistent unified modeling, and we believe that addressing model capacity and data quality will further unlock the full potential of our framework.